\documentclass[journal]{IEEEtran}

\ifCLASSINFOpdf
\else
\fi

\usepackage{cite}
\usepackage{times}
\usepackage{algorithmic}
\usepackage{array}
\usepackage{amssymb}
\usepackage{mdwmath}
\usepackage{mdwtab}
\usepackage{eqparbox}
\usepackage{cite}
\usepackage{url}
\usepackage{multicol,multirow}
\usepackage[cmex10]{amsmath}
\usepackage{makeidx}
\usepackage{diagbox}
\usepackage{rotating,soul}
\usepackage{wrapfig}
\usepackage{booktabs}
\usepackage[usenames,dvipsnames]{color}
\usepackage{ragged2e}
\usepackage{pifont}
\usepackage{overpic}
\makeindex

\usepackage{geometry}
\RequirePackage{silence}
\DeclareGraphicsExtensions{.pdf,.jpg,.png}

\usepackage[pagebackref=false,breaklinks=true,colorlinks, bookmarks=false]{hyperref}

\newcommand{\tabref}[1]{Tab.~\ref{#1}}

\def\ie{\emph{i.e.}}
\def\eg{\emph{e.g.}}
\def\etal{{\em et al.}}

\newcommand{\addFig}[1]{}
\newcommand{\addFigs}[1]{}

\graphicspath{{./figs/}}

\newlength\savedwidth

\begin{document}

\title{Segment Anything Model-guided Collaborative Learning Network for Scribble-supervised Polyp Segmentation}

\author{
Yiming Zhao, 
Tao Zhou,
Yunqi Gu,
Yi Zhou,
Yizhe Zhang,
Ye Wu, 
Huazhu Fu,~\IEEEmembership{Senior Member,~IEEE}
\thanks{Y. Zhao, T. Zhou, Y. Gu, Y. Zhang, and Y. Wu are with PCA Lab, and the School of Computer Science and Engineering, Nanjing University of Science and Technology, Nanjing 210094, China.}
\thanks{Y. Zhou is with the School of Computer Science and Engineering, Southeast University, Nanjing 211189, China.}
\thanks{H. Fu is with the Institute of High Performance Computing, A*STAR, Singapore.}
\thanks{Corresponding author: \textit{Tao Zhou} (taozhou.ai@gmail.com).}
}

\maketitle

\begin{abstract}
Polyp segmentation plays a vital role in accurately locating polyps at an early stage, which holds significant clinical importance for the prevention of colorectal cancer. Various polyp segmentation methods have been developed using fully-supervised deep learning techniques. However, pixel-wise annotation for polyp images by physicians during the diagnosis is both time-consuming and expensive. Moreover, visual foundation models such as the Segment Anything Model (SAM) have shown remarkable performance. Nevertheless, directly applying SAM to medical segmentation may not produce satisfactory results due to the inherent absence of medical knowledge.
In this paper, we propose a novel SAM-guided Collaborative Learning Network (SAM-CLNet) for scribble-supervised polyp segmentation, enabling a collaborative learning process between our segmentation network and SAM to boost the model performance. Specifically, we first propose a Cross-level Enhancement and Aggregation Network (CEA-Net) for weakly-supervised polyp segmentation. Within CEA-Net, we propose a Cross-level Enhancement Module (CEM) that integrates the adjacent features to enhance the representation capabilities of different resolution features. Additionally, a Feature Aggregation Module (FAM) is employed to capture richer features across multiple levels. Moreover, we present a box-augmentation strategy that combines the segmentation maps generated by CEA-Net with scribble annotations to create more precise prompts. These prompts are then fed into SAM, generating segmentation SAM-guided masks, which can provide additional supervision to train CEA-Net effectively. Furthermore, we present an Image-level Filtering Mechanism to filter out unreliable SAM-guided masks. Extensive experimental results show that our SAM-CLNet outperforms state-of-the-art weakly-supervised segmentation methods.
\end{abstract}


\IEEEpeerreviewmaketitle

\begin{figure}[!t]
	\centering
	\footnotesize
	\begin{overpic}[width=1\columnwidth] {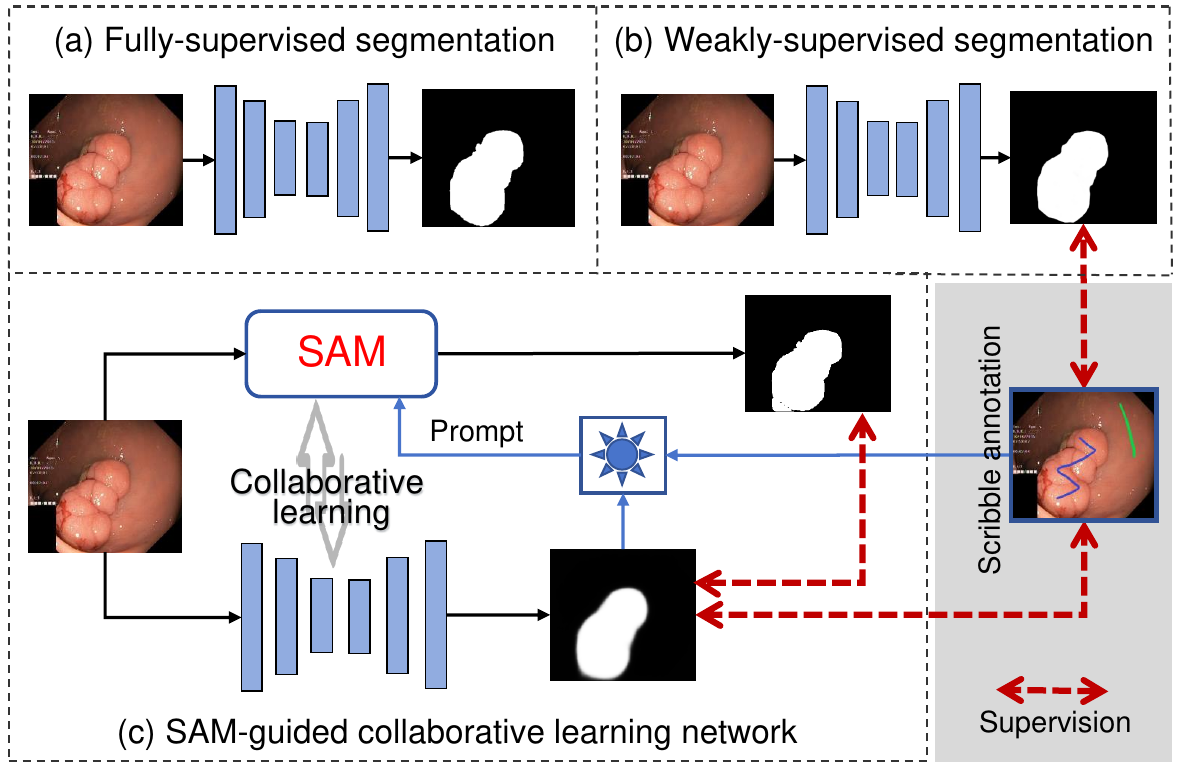}
	\end{overpic}
	\caption{\footnotesize Visual comparison of three different segmentation frameworks: (a) fully-supervised polyp segmentation, (b) weakly-supervised polyp segmentation, and (c) our SAM-guided collaborative learning network for scribble-supervised polyp segmentation.}\vspace{-0.1cm}
	\label{fig1}
\end{figure}

\section{Introduction}

\IEEEPARstart{C}{olorectal} cancer (CRC) stands as one of the most prevalent types of cancer, characterized by a high incidence rate. Its early manifestations often appear in the form of colorectal adenoma polyps~\cite{center2009worldwide}. In clinical practice, colonoscopy serves as the gold standard for identifying diseased tissues within the gastrointestinal tract. Timely identification of colon polyps allows for their prompt removal, thereby preventing further damage to surrounding tissues and reducing the incidence of CRC. In the initial evaluation stage, clinicians often follow a popular procedure of identifying adenomatous polyps, and then these polyps are manually delineated. However, this manual process of detecting and segmenting polyps is time-consuming and subjective. Therefore, the development of a precise and automated polyp segmentation algorithm using colonoscopy images holds paramount clinical significance. 

Deep learning-based polyp segmentation methods have been widely developed and exhibited excellent performance~\cite{fan2020pranet,zhao2021automatic,zhang2021transfuse,dong2023PolypPVT}. Despite the remarkable progress achieved by these methods, it is important to note that they primarily rely on fully-supervised learning (seen in Fig.~\ref{fig1}(a)), which necessitates a substantial number of pixel-wise annotations for training. 
However, collecting annotations for medical image segmentation is an expensive and time-consuming task, as it requires domain-specific experts to annotate at the pixel level. 
To address this challenge, as shown in Fig.~\ref{fig1}(b), weak supervision methods~\cite{ren2023towards,dong2019semantic,zhu2023feddm} have emerged as effective solutions in dealing with limited annotated data and sparse labels. For instance, Ren~\etal~\cite{ren2023towards} proposed a weakly- and semi-supervised method for polyp segmentation. This method trains the segmentation model using only weakly annotated images along with unlabeled images, thereby reducing the reliance on fully annotated data. Dong~\etal~\cite{dong2019semantic} presented a semantic lesions representation transfer model for weakly-supervised endoscopic lesions segmentation, which leverages knowledge from relevant fully-labeled disease segmentation tasks to enhance the performance of target weakly-labeled lesions segmentation tasks. However, polyps often exhibit variations in appearance, such as differences in size, color, and texture, even within the same category. Additionally, the boundary between polyps and surrounding tissue is often blurred, resulting in low contrast. Consequently, the presence of only sparse annotations still poses a significant challenge for polyp segmentation.

In this paper, as shown in Fig.~\ref{fig1}(c), we propose SAM-CLNet, a Collaborative Learning Network guided on the Segment Anything Model (SAM), for scribble-supervised polyp segmentation. SAM-CLNet effectively integrates cross-level features and utilizes masks generated by SAM to enhance model performance. 
Specifically, we first propose a Cross-level Enhancement and Aggregation Network (CEA-Net), in which a Cross-level Enhancement Module (CEM) is proposed to integrate the adjacent features for enhancing the representation capabilities of different resolution features. Additionally, a Feature Aggregation Module (FAM) is proposed to combine encoder features and provide complementary information across multiple levels. 
Moreover, we develop a SAM-guided mask generation network that takes SAM's ability to produce segmentation masks. These masks are then utilized as supplementary supervision labels to train CEA-Net.
To enhance the accuracy of our model, we propose a box-augmentation strategy that combines segmentation maps generated by CEA-Net with scribble annotations to generate more accurate prompts. These prompts are then inputted into SAM, thus we obtain SAM-guided masks. Further, we design an image-level filtering mechanism to eliminate unreliable SAM-guided masks, enabling the quality of the generated masks used for training. 
Finally, we formulate a collaborative learning framework by training CEA-Net and fine-tuning SAM simultaneously. This collaborative approach enhances the interaction between the two components and leads to improved segmentation results.

The main contributions of this paper are listed as follows:
\begin{itemize}
\item We propose a novel SAM-guided CLNet framework for scribble-supervised polyp segmentation, facilitating a collaborative learning process between our segmentation network and SAM. This collaborative approach provides reliable guidance for model training.
\item We present a simple yet effective CEA-Net for achieving weakly-supervised polyp segmentation. In the CEA-Net, a cross-level enhancement module is proposed to integrate adjacent features, while a feature aggregation module cascades multi-level features to provide rich feature representations. These two key modules significantly contribute to the final segmentation performance.
\item Leveraging a box prompt-aided SAM, we generate masks that serve as additional labels for training our segmentation model. To achieve more precise prompt boxes, we design a box-augmentation strategy. Furthermore, we present an image-level filtering mechanism to effectively filter out unreliable masks.

\item We have constructed new scribble-annotated colonoscopy datasets, which are released for further exploration in the domain of weakly-supervised polyp segmentation. Experimental results demonstrate that our model significantly outperforms other state-of-the-art weakly-supervised segmentation methods. 

\end{itemize}

\section{Related Work}
\label{related}

This section reviews several related works, including 1) polyp segmentation, 2) weakly-supervised medical image segmentation, and 3) segment anything model.

\begin{figure*}[t!]
	\centering
        \footnotesize
	\begin{overpic}[width=1.0\textwidth]{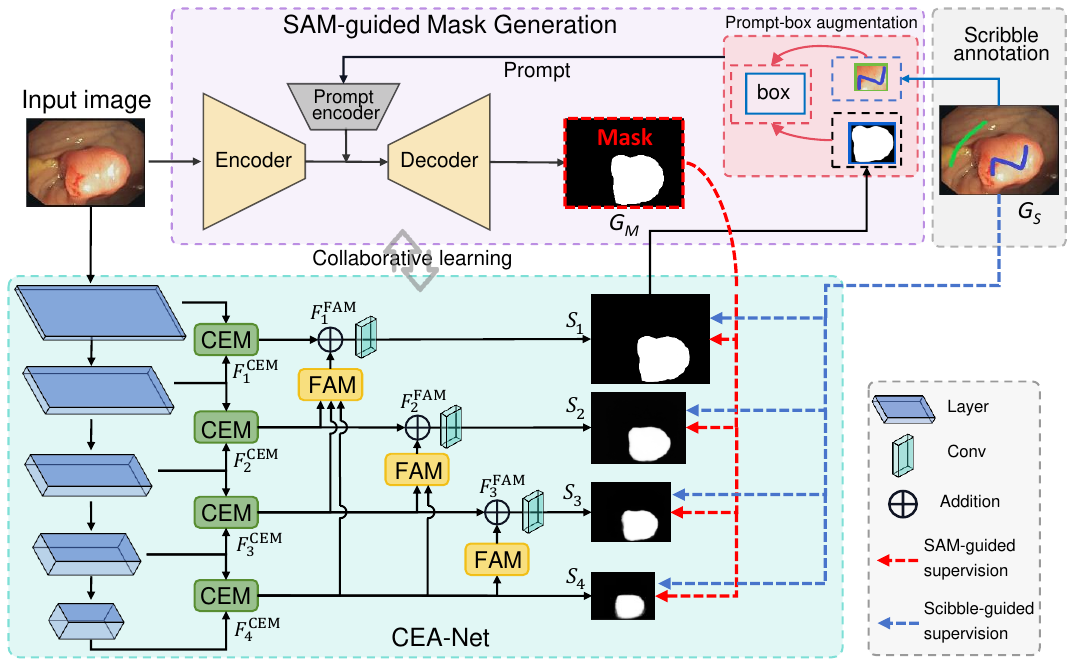}
        \end{overpic}\vspace{-0.25cm}
	\caption{\footnotesize Overall architecture of the proposed SAM-guided CLNet with two key components, \ie, CEA-Net and SAM-guided mask generation network. Within CEA-Net, we propose CEM and FAM to make better use of cross-level features and aggregate them for generating segmentation maps. We combine segmentation maps obtained from CEA-Net and scribble annotations to form prompt boxes, which are passed through SAM to generate SAM-guided masks to provide additional supervision signals.}
    \label{fig2}\vspace{-0.35cm}
\end{figure*}

\subsection{Polyp Segmentation}

With the advancement of deep learning technology, numerous Convolutional neural networks (CNNs)-based methods have been applied to polyp segmentation.
As an exemplary architecture, UNet~\cite{ronneberger2015u} employs an encoder-decoder structure, utilizing skip-connections to incorporate features from the encoder into the decoder. This design enables the comprehensive utilization of semantic information and spatial details from different levels of features. Variants of U-Net, such as UNet++~\cite{zhou2019unet++} and ResUNet++~\cite{jha2021comprehensive}, have also demonstrated advancements in polyp segmentation. To fully exploit contextual information, ACSNet~\cite{zhang2020adaptive} adaptively selects and integrates both global contexts and local information, achieving more robust polyp segmentation performance. CCBANet~\cite{nguyen2021ccbanet} proposes cascading context and attention balance modules to aggregate better feature representations. Additionally, several methods exploit valuable boundary information to improve segmentation accuracy. For instance, Zhou \etal~\cite{zhou2023cross} proposed a cross-level feature aggregate network that simultaneously utilizes boundary details and captures hierarchical semantic information for polyp segmentation. Fang \etal~\cite{fang2019selective} designed a boundary-sensitive loss to introduce area-boundary constraints for generating more precise predictions. Moreover, to deal with the scale variations, several methods~\cite{wu2023meta,song2022attention,zhang2022hsnet} are employed to aggregate multi-scale information for enhancing the ability of feature representation. Furthermore, video polyp segmentation~\cite{ji2022video,jiang2023yona} has recently gained increasing attention.

\subsection{Weakly-supervised Medical Image Segmentation}

Compared to fully-supervised methods, weakly-supervised models aim to reduce their dependency on pixel-level ground-truth labels. In recent years, there have been remarkable advancements in weakly-supervised medical segmentation methods, demonstrating impressive performance. Some of these methods explore the utilization of image-level labels to segment lesion regions. For instance, Wu~\etal~\cite{wu2019weakly} proposed a novel attentional class activation map (CAM) technique that effectively localizes lesion regions using only image-level labels. Chen~\etal~\cite{chen2022c} introduced the causal CAM approach to address challenges in distinguishing the boundary between foreground and background, as well as dealing with pronounced co-occurrence phenomena during training. In addition to image-level labels, several approaches leverage scribbles as annotations to train segmentation networks. For example, Yu~\etal~\cite{yu2021structure} proposed a weakly-supervised cell segmentation framework based on scribble-level annotations. Liu~\etal~\cite{liu2022weakly} presented an effective method for CT slice segmentation of COVID-19 infections, incorporating scribble supervision. The utilization of minimal user interaction has been advocated in some approaches for weakly-supervised segmentation. Roth~\etal~\cite{roth2021going} proposed the use of extreme point clicks as a form of minimal user interaction for training deep learning-based segmentation models. Further, bounding boxes~\cite{kervadec2020bounding,wei2023weakpolyp} have emerged as another commonly used annotation type in weakly-supervised segmentation. Wei~\etal~\cite{wei2023weakpolyp} introduced the mask-to-box (M2B) transformation specifically for polyp segmentation, which relies solely on bounding box annotations.

\subsection{Segment Anything Model}

Segment Anything Model (SAM)~\cite{kirillov2023segment} is a cutting-edge vision foundation model developed for general object segmentation. SAM can generate highly precise segmentation mask results and has garnered immense praise, with some even proclaiming that SAM has revolutionized the field by surpassing existing segmentation models. 
Chen~\etal~\cite{chen2023sam} proposed a SAM-Adapter that merges domain-specific information or visual cues into a segmented network by using simple and efficient adapters. Based on a semi-supervised approach, Zhang~\etal~\cite{zhang2023samdsk} proposed an iterative new method that combines SAM with domain-specific knowledge to reliably build medical image segmentation models from unlabeled images. Additionally, Risab Biswas~\etal~\cite{biswas2023polyp} proposed a text cue-assisted SAM method ({Polyp-SAM++}) to better utilize SAM using text cued for polyp segmentation. Moreover, SAM has been applied to weakly-supervised segmentation tasks~\cite{jiang2023segment,sun2023alternative,chen2023weakly,heC2023weakly,chen2023segment}. For instance, Jiang \etal~\cite{jiang2023segment} proposed to leverage SAM to generate pseudo labels, which are then used to train the segmentation models for weakly-supervised semantic segmentation. He~\etal~\cite{heC2023weakly} utilized the provided sparse annotations as prompts to generate segmentation masks that are adopted to train a concealed object segmentation model. Chen~\etal~\cite{chen2023segment} took advantage of SAM's class-agnostic capability to produce fine-grained instance masks, and then utilized CAM pseudo-labels as cues to select and combine SAM masks, resulting in high-quality pseudo-labels. In contrast to existing works~\cite{jiang2023segment,heC2023weakly}, we aim to develop a collaborative learning framework that can dynamically fine-tune the SAM online, thereby providing more precise masks as additional supervision signals.

\begin{figure*}[t!]
	\centering
	\footnotesize
	\begin{overpic}[width=1\textwidth]{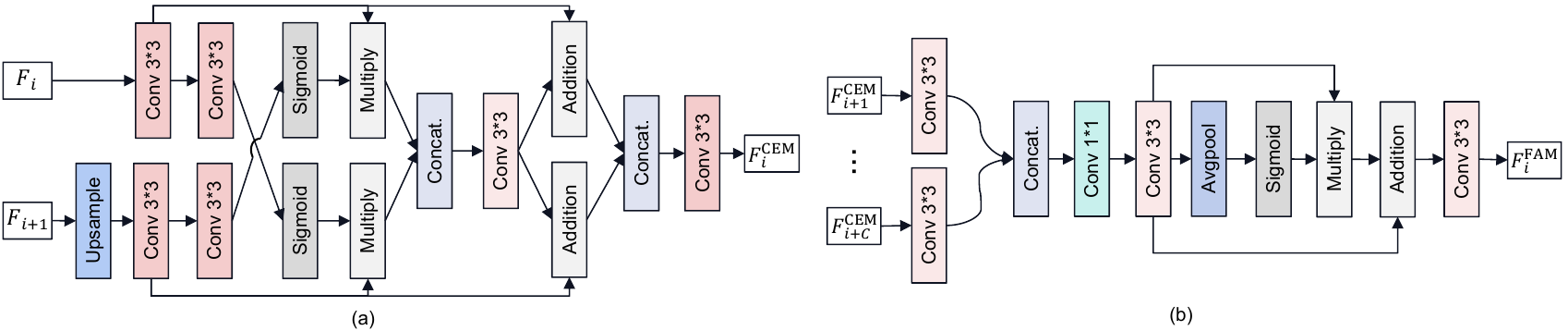}
	\end{overpic}\vspace{-0.15cm}
	\caption{\footnotesize Illustrations of the proposed modules: (a) cross-level enhancement module and (b) feature aggregation module.}
	\label{fig:mod}\vspace{-0.25cm}
\end{figure*}

\section{Proposed Method}

In this section, we first delineate the overall architecture of the proposed SAM-CLNet. Then we introduce our cross-level enhancement and aggregation network, and provide the details of SAM-guided mask generation. Finally, we present the comprehensive loss function of our framework.

\subsection{Overview}

In this study, we propose a novel SAM-CLNet framework for weakly-supervised polyp segmentation, aiming to learn a segmentation model from sparsely annotated datasets. As shown in Fig.~\ref{fig2}, the proposed SAM-CLNet consists of two essential components, namely Cross-level Enhancement and Aggregation Network (CEA-Net) and SAM-guided Mask Generation Network (SAM-MGNet). In our CEA-Net, an input image $I\in{R^{W\times H\times 3}}$ is first fed to the encoder model (Res2Net-50~\cite{gao2019res2net} as the backbone) to extract multi-level features, denoted as $\{F_i, i=1,2,\dots,5\}$. Then we obtain the feature resolution of $\frac{W}{4}\times\frac{H}{4}$ for the first level and $\frac{W}{2^i}\times\frac{H}{2^i}$ (when $i>1$) as a general resolution. The channel number of features at the $i$-th layer is given as $c_i~(i=1,2,\dots,5)$ with $C=[64, 256, 512, 1024, 2048]$. To fully integrate the multi-level features, we utilize the proposed CEM to capture contextual information from adjacent resolutions. This module employs a cross-enhanced strategy that collaboratively learns knowledge guidance to boost the segmentation performance. 

Additionally, the proposed FAM is used to cascade the integrated cross-level features, resulting in the production of multiple side-out segmentation maps (denoted as $S_{i},i=1,2,3,4$, with $S_1$ being the main segmentation map). Moreover, within the SAM-MGNet, we leverage the main segmentation map $S_1$ along with scribble annotations to generate prompts. These prompts are then adapted to generate masks, serving as additional labels for training our CEA-Net. As a result, our CEA-Net and SAM-MGNet can be collaboratively optimized to achieve more promising segmentation results. The details of key components will be presented in subsequent sections.

\subsection{Cross-level Enhancement and Aggregation Network }

In this study, we propose a CEA-Net to fully capture the contextual semantic information and make better use of multi-level features. The CEA-Net is regarded as the segmentation model to generate predicted segmentation maps, which are supervised by using scribble annotations and SAM-guided masks. Within CEA-Net, we propose a CEM to fuse the cross-level features, which can effectively integrate the contextual information from adjacent resolutions. Additionally, we propose FAM to aggregate the enhanced features from each CEM to capture richer feature representations. Both the two modules can collaboratively contribute to the final segmentation performance.

\subsubsection{Cross-level Enhancement Module}

In the feature extraction network, convolutional layers at different levels are responsible for capturing features at different levels of abstraction. The integration of these multi-level features plays a vital role in enhancing the model's representation capability across various resolutions \cite{pang2020multi}. To improve the effectiveness of this integration process, we propose the CEM to enhance features with different resolutions from adjacent layers in a cross-enhanced manner. As shown in Fig.~\ref{fig:mod}(a), taking $F_{i}$ and $F_{i+1}$ as the two adjacent features, the high-level feature $F_{i+1}$ is adjusted through an upsample operation to have the same resolution with $F_{i}$. Then, both features are passed through a $3\times{3}$ convolutional layer, resulting in preliminary features $F_l =Conv_{3\times{3}}(F_i)$ and $F_h=Conv_{3\times{3}}(\mathcal{U}(F_{i+1})))$. Here, $F_h$ and $F_l$ respectively represent the lower and higher layers of the adjacent features, respectively, and $\mathcal{U}(\cdot)$ denotes an up-sampling operation. 
Subsequently, the two features are fed into a $3\times{3}$ convolutional layer and then passed through a Sigmoid function, which can normalize values to a range of $[0,1]$. By doing such a process, we obtain two weight maps that can be interpreted as feature-level spatial attention maps. These attention maps facilitate the mutual enhancement of the two features at the feature level. By utilizing this cross-enhanced strategy, we can obtain the enhanced features, which are formulated as follows:
\begin{equation} 
    \left\{
        \begin{aligned}
        & F^{en}_l=\mathcal{S}(Conv_{3\times{3}}(F_h))\odot F_l,\\
        & F^{en}_h=\mathcal{S}(Conv_{3\times{3}}(F_l))\odot F_h,
        \end{aligned}\label{eq13}
    \right.
\end{equation}
where $\mathcal{S}(\cdot)$ is the Sigmoid activation function, and $\odot$ represents the element-wise multiplication. Furthermore, we aggregate the two cross-enhanced features, and the cascaded feature is then passed through a convolutional block ($Bconv_{3\times{3}}$), including $3\times{3}$ convolution, followed by batch normalization and ReLU activation function. As a result, we obtain the final cascaded feature $F^{cat}$ by 
\begin{equation}
\label{equ:w_all}
F^{cat}=Bconv_{3\times{3}}(\mathcal{O}_{cat}(F^{en}_l,F^{en}_h)),
\end{equation}
where $\mathcal{O}_{cat}(\cdot)$ represents the concatenation operation. To preserve the original feature information from each layer, we apply a residual connection by stacking $F_l$ and $F_h$ with the initial enhanced features $F^{en}_l$ and $F^{en}_h$, respectively. This ensures that the model retains the original features while incorporating the cross-enhanced information. Ultimately, the final cross-level fused feature can be represented as follows:
\begin{equation}
\label{eq3}
F_i^{\textup{CEM}}=Conv_{3\times{3}}(\mathcal{O}_{cat}(F_l\oplus{F_{cat}},F_h\oplus{F_{cat}})),
\end{equation}
where $\oplus$ represents element-wise addition, and $F_i^{\textup{CEM}}$ denotes the oupput of the $i$-th CEM. The CEM incorporates a cross-enhanced mechanism that enables $F_i^{\textup{CEM}}$ to contain rich and complementary information from cross-level features. This enhances the feature representation capability of the model. By employing the CEM at each pair of adjacent layers, we obtain outputs for each pair, \ie, $F_i^{\textup{CEM}}$ ($i=1,2,3,4$).


\subsubsection{Feature Aggregation Module}

It is important to incorporate useful information from the encoder into the decoder, resulting in improved segmentation performance. To capture much richer feature representations, we devise a multi-level feature aggregation module to make better use of all available features from different levels in the encoder, as shown in Fig.~\ref{fig:mod}(b). Considering the potential dilution of certain intricate features, we extract the features from their corresponding layers as well as the preceding layers' output features, thereby anticipating polyp regions. Specifically, when incorporating the feature $F_i^{\textup{CEM}}$ at $i$-the layer into the decoder, we first aggregate all available features from previous CEMs (\eg, $F_{i+1}^{\textup{CEM}}$, $F_{i+2}^{\textup{CEM}}$, etc.) and combine the aggregated feature with $F_i^{\textup{CEM}}$, which is then fed into the decoder to generate the predicted segmentation map. In this case, all the previous features are cascaded and then passed through a 1$\times$1 convolution to reduce the channel. Thus, we can obtain the cascaded feature $F^{cas}$, which can be formulated by
\begin{equation} 
\begin{aligned}
F^{cas}=Conv_{1\times{1}}(\mathcal{O}_{cat}(Bconv_{3\times{3}}(F^{\textup{CEM}}_{i+1}),\\Bconv_{3\times{3}}(F^{\textup{CEM}}_{i+2}),...,)). \\
\end{aligned}
\end{equation}

Note that the FAM can dynamically accommodate different numbers of inputs. The feature $F^{cas}$ is passed through a $3\times{3}$ convolutional block, resulting in the feature $F^{con}$. Then, we employ a global average pooling operation on $F^{con}$, which is further passed through a Sigmoid activation function to generate the global weight map. The weight map is element-wise multiplied with the corresponding feature $F^{con}$, producing the global attention-enhanced feature $F^{g}$. Then, to preserve the original feature information from each layer, we apply a residual connection by stacking $F^{con}$ and $F^{g}$ with the initial enhanced features. After that, the fused feature is passed through a $3\times{3}$ convolutional block, to obtain the final aggregated feature. The above process can be represented by
\begin{equation}
F_i^{\textup{FAM}}=Bconv_{3\times{3}}((F^{con}\odot \mathcal{S}(\mathcal{O}_{avg}(F^{con}))) \oplus{F^{con}}),
\end{equation}
where $\mathcal{O}_{avg}$ represents the global average pooling operation. 

Sequentially, the feature $F_{i}^{\textup{CEM}}$ is combined with $F_i^{\textup{FAM}}$ by using an addition operation, and the fused feature is passed through a $3\times{3}$ convolution block, resulting in the feature $F_{i}^{fuse}$. Finally, $F_{i}^{fuse}$ is passed through a $1\times{1}$ convolutional layer to generate segmentation maps $S_{i}$ $(i=1,2,3,4)$. 

\begin{table*}[t!]
  \centering
  \small
  \renewcommand{\arraystretch}{1.1}
  \setlength\tabcolsep{5.0pt}
  \caption{Quantitative results on the {Kvasir} and {CVC-ClinicDB} datasets. The first two best results are marked in red and blue. }\label{tab:2result}
  \vspace{-0.2cm}
  \begin{tabular}{c|r||cccccc|cccccc}
  \toprule
  \multicolumn{2}{c||}{\multirow{2}*{\textbf{Methods}}} & \multicolumn{6}{c|}{Kvasir} & \multicolumn{6}{c}{CVC-ClinicDB} \\
  \cline{3-14}
  \multicolumn{2}{c||}{} & mDice & mIoU  & $S_{\alpha}$  &  $F_\beta^w$ &  $E_\phi^{max}$  & MAE & mDice & mIoU  & $S_{\alpha}$  &  $F_\beta^w$ & $E_\phi^{max}$ & MAE\\
  \midrule
 
  \multirow{6}{*}{\begin{sideways}Fully-supervised \end{sideways}} 

  & UNet++\cite{zhou2019unet++}  & 0.821 & 0.744 & 0.862 & 0.808 & 0.910 & 0.048 & 0.794 & 0.729 & 0.873 & 0.785 & 0.931 & 0.022\\
  
  & ACSNet\cite{zhang2020adaptive}  & 0.898 & 0.838 & 0.920 & 0.882 & 0.952 & 0.032 & 0.882 & 0.826 & 0.927 & 0.873 & 0.959 & 0.011\\
  
  & EU-Net\cite{patel2021enhanced}  & 0.908 & 0.854 & 0.917 & 0.893 & 0.954 & 0.028 & 0.902 & 0.846 & 0.936 & 0.891 & 0.965 & 0.011\\
  & MSEG\cite{huang2021hardnet}  &0.897 &0.839 & 0.912 &0.885 &0.948 &0.028 & 0.909 & 0.864 & 0.938 & 0.907 &0.969  &0.007\\
  & MSNet\cite{zhao2021automatic} &0.905 &0.849 & 0.923 &0.892 &0.954 &0.028 & 0.918 & 0.869 & 0.946 &0.913 &0.979 &0.008 \\
  & DCRNet\cite{yin2022duplex}  & 0.886 & 0.825 & 0.911 & 0.868 & 0.941 & 0.035 & 0.896 & 0.844 & 0.933 & 0.890 & 0.978 & 0.010\\
  
  \midrule

  \multirow{8}{*}{\begin{sideways}Weakly-supervised  
  \end{sideways}} 
  & {C2FNet}$^{\dagger}$\cite{sun2021context}& 0.819 & 0.732 & 0.862 & 0.815 & 0.908 & 0.050 & 0.776 & 0.690 & 0.853 & 0.770 & 0.924 & 0.026\\
  & PraNet$^{\dagger}$\cite{fan2020pranet} &  0.793 & 0.704 & 0.850 & 0.778 & 0.909 & 0.051 
  & 0.715 & 0.620 & 0.821 & 0.697 & 0.906 & 0.034\\

  \cline{2-14}
  & WSSOD\cite{zhang2020weakly}  & 0.781 & 0.679 & 0.834 & 0.767 & 0.909 & 0.054 & 0.765 & 0.659 & 0.848 & 0.771 & 0.933 & {0.022}\\
  & SCWSSOD\cite{yu2021structure} & 0.764 & 0.671 & 0.823 & 0.768 & 0.874 & 0.058 & 0.726 & 0.631 & 0.816 & 0.737 & 0.884 & 0.035\\
  & WSCOD\cite{he2023weakly} & 0.805 & 0.719 & 0.847 & 0.788 & 0.899 & {0.058} & 0.742 & 0.649 & 0.822 & 0.738 & 0.907 & {0.037}\\
  & SBANet\cite{huang2022scribble} & 0.753 & 0.658 & 0.818 & 0.681 & 0.861 & 0.084 & 0.739 & 0.645 & 0.841 & 0.658 & 0.907 & 0.044\\
  \cline{2-14}
  & CEA-Net (Ours) & \textcolor{blue}{0.842} & \textcolor{blue}{0.769} & \textcolor{blue}{0.881} & \textcolor{blue}{0.826} & \textcolor{blue}{0.917} & \textcolor{blue}{0.045} & \textcolor{blue}{0.845} & \textcolor{blue}{0.781} & \textcolor{blue}{0.891} & \textcolor{blue}{0.828} & \textcolor{blue}{0.947} & \textcolor{blue}{0.021}\\
  & SAM-CLNet (Ours) & \textcolor{red}{0.857} & \textcolor{red}{0.788} & \textcolor{red}{0.882} & \textcolor{red}{0.854} & \textcolor{red}{0.929} & \textcolor{red}{0.042} & \textcolor{red}{0.853} & \textcolor{red}{0.783} & \textcolor{red}{0.897} & \textcolor{red}{0.845} & \textcolor{red}{0.953} & \textcolor{red}{0.014}\\
  \bottomrule
  \end{tabular}\vspace{-0.15cm}
\end{table*}

\subsection{SAM-guided Mask Generation}

Recently, SAM has demonstrated significant effectiveness in numerous segmentation tasks within the realm of computer vision. However, existing studies~\cite{zhou2023can,li2023polyp} show that SAM faces substantial challenges when applied to the field of medical image segmentation, particularly for polyp segmentation. Two primary factors contribute to these difficulties. First, SAM relies on ``prompts" to provide clues about the object of interest and generate segmentation results. Yet, these prompts currently need to be manually input, which is a significant limitation. Second, SAM often struggles to accurately segment target polyps in images where boundary information is unclear and when there is a high degree of similarity between the foreground target and background, even when provided with the right ``prompts".
To address these challenges, We propose a collaborative learning framework, wherein our CEA-Net assumes the role of generating segmentation maps. These maps are then combined with scribble annotations to create precise prompts. Subsequently, these prompts are processed by SAM to generate SAM-guided masks which serve as supplementary supervision signals. As a result, our CEA-Net and SAM engage in a collaborative learning process, allowing both components to effectively work together and achieve improved accuracy in segmentation performance.

\subsubsection{Prompt Generation}

\begin{figure}[!t]
  \centering
  \footnotesize
  \includegraphics[width=0.5\textwidth]{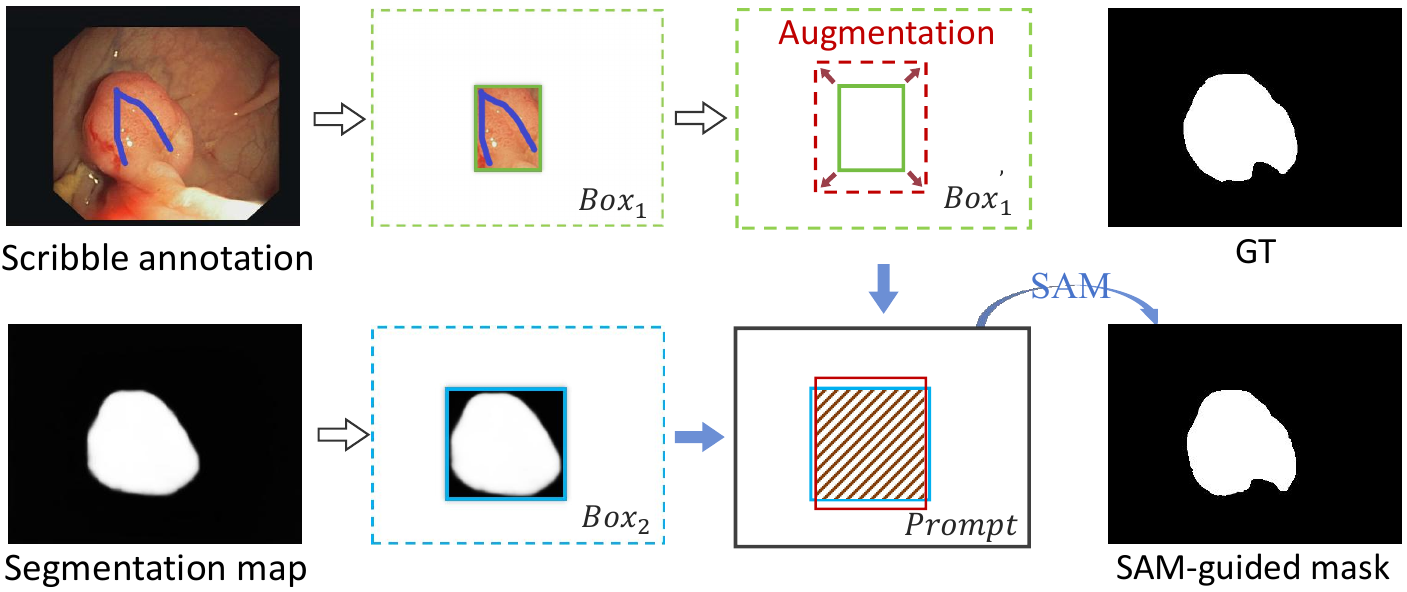}
  \caption{\footnotesize Process of prompt generation, where the blue one is from segmentation map with our CEA-Net, the green box is generated by a scribble annotation, and the red one is augmentation box based on the green box. The prompt is formed by utilizing the overlapping region between the red and blue boxes.}
  \label{box}\vspace{-0.1cm}
\end{figure}

We devise an effective strategy for generating the prompt by combining scribble annotations with segmentation maps obtained from our CEA-Net. Since our label consists of line-based scribble annotations, which contain fewer pixels compared to the polyp target that needs to be segmented.  If we solely rely on the scribble annotations to generate prompts, although the accuracy of the annotated area is high, the resulting box (denoted as ``$Box_{1}$") may not fully encompass the polyp target, leading to a loss of valuable target information. To address this issue, we aim to utilize the segmentation maps generated by our CEA-Net to expand the areas not covered by the scribble annotation boxes. However, it is important to note that the box (denoted as ``$Box_{2}$") obtained from the segmentation map may be significantly larger than the predicted segmentation target. While the segmentation map exhibits high accuracy, allowing the segmentation box to expand without limitations would be undesirable. Therefore, as illustrated in Fig.~\ref{box}, we present an augmentation-based approach. We have green and blue rectangular boxes obtained from the scribble annotation and the segmentation map, respectively. We then expand the scribble annotation box by a certain number of pixels, resulting in an augmented box (``$Box^{'}_{1}$" indicated by the red dotted rectangular box in Fig.~\ref{box}). Our prompt box is determined by taking the intersection between the augmented box (``$Box^{'}_{1}$") and the blue rectangular box (``$Box_{2}$"). It is worth noting that this strategy can be regarded as prompt engineering, which strikes a balance by expanding the prompt box to include missed regions while preventing excessive expansion that could encompass unrelated areas. By applying this strategy, we can generate more precise prompts that effectively capture the complete polyp target while minimizing the inclusion of irrelevant regions.

\subsubsection{Image-level Filtering Mechanism}

Our study reveals that SAM tends to produce both inaccurate and seemingly reasonable segmentation results while handling polyp images with indistinct edge information. These inaccurate results can unfortunately lead to the generation of misleading supervisory signals. To address this issue, we devise a filtering mechanism that screens SAM-guided masks and discards inaccurate segmentations during supervised learning. More specifically, we evaluate the similarity between SAM-guided masks and scribble annotations, setting a similarity threshold $\tau$ (\ie, $0.5$ in this study) to filter out unreliable SAM-guided masks. If $\tau<0.5$, the corresponding SAM-guided mask is discarded during the computation of difference loss between the SAM-guided masks and segmentation maps derived from our own CEA-Net. To help facilitate this, we use an indicator vector $o$. In this vector, $o_i=1$ indicates that the $i$-th SAM-guided mask is reliable, and $o_i=0$ denotes an unreliable SAM-guided mask. Through this mechanism, we aim to provide reliable supervision for training an accurate segmentation model by selectively filtering out inaccurate SAM-guided masks.

\subsection{Overall Objective Function}

\subsubsection{Partial Cross-entropy Loss}
We employ partial cross-entropy loss to quantify the disparities between the segmentation results and scribble annotations, which is defined by
\begin{equation}
  \mathcal{L}_{pce}(\hat{y},y)=\frac{1}{N}\sum_{i\epsilon \widetilde{S}}^N -y_{i}log(\hat{y}_i)-(1-y_{i})log(1-\hat{y_i}),
\end{equation}
where $\widetilde{S}$ is the set of labeled pixels in scribble annotations that also encompass both foreground and background, $y_i$ is the true class of pixel $i$ while $\hat{y_i}$ is the prediction on pixel $i$.

\subsubsection{Structure Consistency Loss}

Inspired by~\cite{yu2021structure}, a robust segmentation model should possess the capability to accurately identify the same targets within images, even in the presence of a series of transformations. For a neural network ($F_{\theta}(\cdot)$), characterized by parameters $\theta$, in conjunction with input $x$ and transformation $\mathcal{T}(\cdot)$, the anticipated outcome can be delineated by $F_{\theta}(\mathcal{T}(x))=\mathcal{T}(F_{\theta}(x))$. To enhance the stability of our model, we introduce a structure consistency loss to minimize the difference between the segmentation results under different scale inputs. Specifically, we resize input images $X$ to reduce it to $0.3$ of the original size to obtain $X^{\downarrow}$, and then put it into our segmentation network to obtain the downsized prediction map $S_1^{\downarrow}$. Note that, the structure consistency loss is only computed at the segmentation map of the first layer (\ie, $S_1$). Thus, our structure consistency loss can be formulated as follows:
\begin{equation}
\mathcal{L}_{ss}(S_1,S^{\downarrow})=\frac{1}{N}\sum_{j=1}^N\|F_{\theta}(\mathcal{T}(x_j^{\downarrow}))-\mathcal{T}(F_{\theta}(x_j))\|_2^2.
\end{equation}

\subsubsection{Weighted Loss of BCE and IoU}

Inspired by~\cite{wei2020f3net}, our segmentation loss, to measure the difference between the SAM-guided masks and generated maps obtained by our CEA-Net, is defined as $\mathcal{L}_{\textup{seg}}=\mathcal{L}_{\textup{wIoU}}+\mathcal{L}_{\textup{wBCE}}$. Here, $\mathcal{L}_{\textup{wIoU}}$ and $\mathcal{L}_{\textup{wBCE}}$ represent the weighted intersection-over-union (wIoU) loss and binary cross-entropy (wBCE) loss~\cite{wei2020f3net}, respectively.

\textbf{Overall Loss:} 
Our loss function combines the dominant loss (regarding $S_1$), and three auxiliary losses (regarding $S_i$ (i=2,3,4)), which can be formulated by 
\begin{equation} 
    \left\{
        \begin{aligned}
        & \mathcal{L}_{dom}=\mathcal{L}_{pce}(S_1,G_S)+\alpha \mathcal{L}_{seg}(o(S_1,G_M))+\mathcal{L}_{ss}(S_1,S^{\downarrow}),\\
        & \mathcal{L}_{aux}^i=\mathcal{L}_{pce}^i(S_i^{\uparrow},G_S)+\alpha  \mathcal{L}_{seg}^i(o(S_i^{\uparrow},G_M)), \quad  i=2,3,4,\\
        \end{aligned}\label{eq13}
    \right.
\end{equation}
where the indicator vector $o$ is used to filter out unreliable SAM-guided masks. $G_S$ and $G_M$ denote scribble annotations and SAM-guided masks, respectively. $S_i^{\uparrow}$ denotes to scale $S_i$ up to have the same size as $S_1$. Besides, $\alpha$ is a trade-off parameter that is set to $0.5$ in this study.

Finally, the overall objective function of our framework is:
\begin{equation}
  \mathcal{L}_{total}=\mathcal{L}_{dom}+\sum_{i=2}^4 \lambda_i \mathcal{L}_{aux}^i,
\end{equation}
where $\lambda_i$ is used to balance the auxiliary loss at each stage. Following in~\cite{yu2021structure}, we set $\lambda_2=0.8$, $\lambda_3=0.6$, and $\lambda_4=0.4$.

\begin{table*}[!t]
  \centering
  \small
  \renewcommand{\arraystretch}{1.2}
  \setlength\tabcolsep{1.5pt}
  \caption{Quantitative results on three unseen datatsets (\ie, CVC-300, CVC-ColonDB, and ETIS-LaribPolypDB). 
  }\label{tab:3result}
  \vspace{-0.2cm}
  \begin{tabular}{c|r||cccccc|cccccc|cccccc}
  \toprule

  \multicolumn{2}{c||}{\multirow{2}*{\textbf{Methods}}} & \multicolumn{6}{c|}{CVC-300}& \multicolumn{6}{c|}{ETIS-LaribPolypDB} & \multicolumn{6}{c}{CVC-ColonDB} \\
  \cline{3-20}
  
  \multicolumn{2}{c||}{} & mDice & mIoU  & $S_{\alpha}$  &  $F_\beta^w$ &  $E_\phi^{max}$  & MAE & mDice & mIoU  & $S_{\alpha}$  &  $F_\beta^w$ & $E_\phi^{max}$ & MAE
  & mDice & mIoU  & $S_{\alpha}$  &  $F_\beta^w$ &  $E_\phi^{max}$  & MAE \\
  \midrule
 
  \multirow{6}{*}{\begin{sideways}Fully-supervised \end{sideways}} 
  & UNet++\cite{zhou2019unet++} & 0.707 & 0.624 & 0.839 & 0.687 & 0.898 & 0.018   
  & 0.401 & 0.344 & 0.683 & 0.390 & 0.776 & 0.035
  & 0.482 & 0.408 & 0.693 & 0.467 & 0.764 & 0.061\\
  
   & ACSNet\cite{zhang2020adaptive} & 0.863 & 0.787 & 0.923 & 0.825 & 0.968 & 0.013   
  & 0.578 & 0.509 & 0.754 & 0.530 & 0.764 & 0.059
  & 0.716 & 0.649 & 0.829 & 0.697 & 0.851 & 0.039\\
  & EU-Net\cite{patel2021enhanced} & 0.837 & 0.765 & 0.904 & 0.805 & 0.933 & 0.015 
  & 0.687 & 0.609 & 0.793 & 0.636 & 0.841 & 0.067
  & 0.756 & 0.681 & 0.831 & 0.730 & 0.872 & 0.045\\
  & MSEG\cite{huang2021hardnet}  & 0.874 & 0.804 & 0.924 & 0.852 & 0.957 & 0.009
  & 0.700 & 0.630 & 0.828 & 0.671 & 0.890 & 0.015
  & 0.735 & 0.666 & 0.834 & 0.724 & 0.875 & 0.038 \\
  & MSNet\cite{zhao2021automatic} & 0.865 & 0.799 & 0.926 & 0.848 & 0.953 & 0.010 
  & 0.723 & 0.652 & 0.845 & 0.677 & 0.890 & 0.020
  & 0.751 & 0.671 & 0.838 & 0.736 & 0.883 & 0.041\\
  & DCRNet\cite{yin2022duplex} & 0.856 & 0.788 & 0.921 & 0.830 & 0.960 & 0.010 
  & 0.556 & 0.496 & 0.736 & 0.506 & 0.773 & 0.096 
  & 0.704 & 0.631 & 0.821 & 0.684 & 0.848 & 0.052 \\
  \midrule

  \multirow{8}{*}{\begin{sideways}Weakly-supervised
  \end{sideways}} 
  & C2FNet$^{\dagger}$\cite{sun2021context}  & 0.786 & 0.675 & 0.860 & 0.744 & 0.932 & 0.013 
  & 0.574 & 0.487 & 0.754 & 0.557 & 0.820 & 0.027 
  & 0.644 & 0.540 & 0.762 & \textcolor{blue}{0.625}  & 0.814 & 0.056\\
  & PraNet$^{\dagger}$\cite{fan2020pranet}  & 0.773 & 0.656 & 0.873 & 0.745 & 0.957 & \textcolor{blue}{0.012}
  & 0.468 & 0.377 & 0.714 & 0.420 & 0.804 & 0.033 
  & 0.488 & 0.399 & 0.698 & 0.474 & 0.763 & 0.056\\
  
    \cline{2-20}
  & WSSOD\cite{zhang2020weakly} & 0.741 & 0.612 & 0.841 & 0.732 & 0.937 & 0.015 
  & 0.445 & 0.357 & 0.689 & 0.424 & 0.793 & 0.028 
  & 0.533 & 0.434 & 0.718 & 0.530 & \textcolor{blue}{0.832} & {0.050}\\
  & SCWSSOD\cite{yu2021structure}  & 0.750 & 0.620 & 0.831 & 0.757 & 0.921 & 0.013 
  & 0.546 & 0.458 & 0.732 & 0.546 & 0.818 & {0.024} 
  & 0.526 & 0.429 & 0.685 & 0.527 & 0.749 & 0.088\\
  & WSCOD\cite{he2023weakly}  & 0.798 & 0.687 & 0.869 & 0.797 & 0.951 & 0.013 
  & 0.484 & 0.409 & 0.700 & {0.470} & 0.774 & 0.058 
  & 0.570 & 0.480 & 0.727 & {0.572} & {0.784} & 0.067\\
  & SBANet\cite{wei2021shallow}  & 0.663 & 0.565 & 0.827 & 0.637 & 0.926 & 0.017 
  & 0.394 & 0.330 & 0.678 & 0.359 & 0.770 & 0.034 
  & 0.474 & 0.394 & 0.705 & 0.458 & 0.794 & 0.059\\
  
  \cline{2-20}
  & CEA-Net  & \textcolor{blue}{0.866} & \textcolor{blue}{0.791} & \textcolor{blue}{0.915} & \textcolor{blue}{0.812} & \textcolor{blue}{0.961} & \textcolor{red}{0.008} 
  & \textcolor{blue}{0.625} & \textcolor{blue}{0.537} & \textcolor{blue}{0.782} & \textcolor{blue}{0.582} & \textcolor{blue}{0.824} & \textcolor{blue}{0.023} 
  & \textcolor{blue}{0.655} & \textcolor{blue}{0.573} & \textcolor{blue}{0.778} & 0.623 & {0.828} & \textcolor{blue}{0.047}\\
  
  & SAM-CLNet & \textcolor{red}{0.876} & \textcolor{red}{0.800} & \textcolor{red}{0.922} & \textcolor{red}{0.851} & \textcolor{red}{0.975} & \textcolor{red}{0.008} 
  & \textcolor{red}{0.665} & \textcolor{red}{0.586} & \textcolor{red}{0.810} & \textcolor{red}{0.635} & \textcolor{red}{0.859} & \textcolor{red}{0.019} 
  & \textcolor{red}{0.711} & \textcolor{red}{0.627} & \textcolor{red}{0.815} & \textcolor{red}{0.694} & \textcolor{red}{0.862} & \textcolor{red}{0.040}\\
  \midrule

  \end{tabular}\vspace{-0.1cm}
\end{table*}

\section{EXPERIMENTS AND RESULTS}\label{sec:Experiments}

In this section, we first present the used datasets in our experiments. Then we present experimental settings, including implementation details and evaluation metrics. Further, we compare the proposed models with the state-of-the-art methods. At last, an ablation study is implemented to investigate the importance of each key component and different strategies in our SAM-CLNet.

\subsection{Datasets}
\subsubsection{Polyp Dataset}
Experiments are conducted on five benchmark colonoscopy datasets, including CVC-ClinicDB\cite{bernal2015wm}, Kvasir\cite{jha2020kvasir}, ETIS-LaribPolypDB\cite{silva2014toward}, CVC-300\cite{vazquez2017benchmark}, and CVC-ColonDB~\cite{tajbakhsh2015automated}. Following the same settings in PraNet\cite{fan2020pranet}, $900$ images were selected from Kvasir and $550$ images from CVC-ClinicDB to form the training set. The remaining images from the two datasets (\ie, Kvasir and CVC-ClinicDB) and the other three datasets (\ie, CVC-ColonDB, ETIS-LaribPolypDB, and CVC-300) are used for testing. 

\subsubsection{Scribble Annotations} 

Due to the absence of weak-annotated colonoscopy datasets, we re-label the $1,450$ images. Following the scribble annotation methodology in~\cite{yu2021structure}, scribble marks are made by drawing simple lines. The foreground polyp region is denoted by a blue line, while the background area is delineated with a green line. Our labeling process is based on the main regions of polyps in the image, without placing undue reliance on the GT entire objects. Remarkably, the task of scribble annotations can be efficiently completed within 3 $\sim$ 4 seconds per image. Note that, we will release the scribble-annotation polyp datasets for further exploration. 

\begin{figure*}[t!]
	\centering
        \footnotesize
	\begin{overpic}[width=1.0\textwidth]{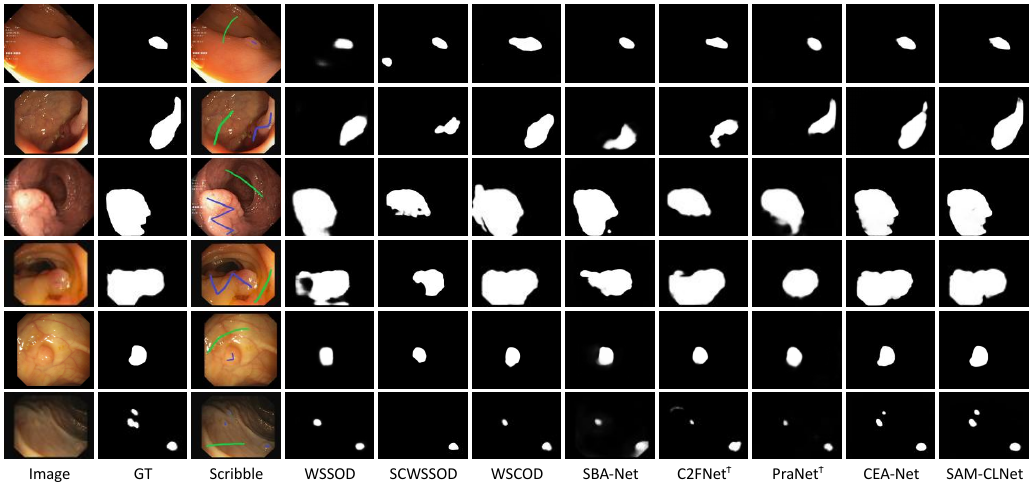}
        \end{overpic}\vspace{-0.2cm}
	\caption{\footnotesize Visual segmentation results obtained by our method and other compared weakly-supervised segmentation methods. 
 }
    \label{fig5}\vspace{-0.15cm}
\end{figure*}

\subsection{Implementation Details and Evaluation Metrics}

\subsubsection{Implementation Details}
Our model is trained using the Stochastic Gradient Descent (SGD) optimizer with a batch size of 12, a momentum of 0.9, and a weight decay of 5e-4. All input images are first resized to 320$\times$320 and then undergo random cropping before being passed through the SAM and our segmentation network. We employ the lightweight {SAM} (\ie, ``vit\_b") featuring a 12-layer encoder, and we freeze the initial 8 layers, unfreezing the final 4 layers to fine-tune SAM to exquisite features, but we freeze the immutability of SAM's decoder and the Prompt encoder to ensure model stability. To optimize the training process, we adopt triangular warm-up and decay strategies~\cite{yu2021structure}, with a maximum learning rate set to 1e-2 and a minimum learning rate of 1e-5. The overall framework is trained over 100 epochs. All experiments are conducted on one NVIDIA GeForce RTX 3090 GPU. In the inference stage, a test image is only passed through our CEA-Net, and the output $S_1$ is taken as the final segmentation map.


\subsubsection{Evaluation Metrics}
To make a more comprehensive quantitative comparison, we adopt six widely used evaluation metrics\cite{zhou2021specificity,fan2021concealed}, namely mean dice score (mDice), mean intersection over union (mIoU), S-measure ($S_\alpha$), F-measure ($F_\beta^\omega$), E-measure ($E_\phi^{max}$), and mean absolute error (MAE).

\subsection{Results and Comparisons}

\subsubsection{Comparison Methods}

To assess the effectiveness of the proposed SAM-CLNet, we conduct a comprehensive comparative analysis with twelve cutting-edge methodologies, including  UNet++\cite{zhou2019unet++}, ACSNet\cite{zhang2020adaptive}, DCRNet\cite{yin2022duplex}, EU-Net\cite{patel2021enhanced}, MSEG\cite{huang2021hardnet}, 
MSNet\cite{zhao2021automatic}, 
C2FNet\cite{sun2021context}, PraNet\cite{fan2020pranet}, WSSOD\cite{zhang2020weakly}, SCWSSOD\cite{yu2021structure}, WSCOD\cite{he2023weakly}, and SBANet\cite{huang2022scribble}. The first eight methods are fully-supervised approaches, while the latter four methods are weakly-supervised methods using scribble annotations. Additionally, we change two fully-supervised methods (\ie, C2FNet and PraNet) into the weakly-supervised setting. For a fair comparison, we change the backbones of these compared methods (\ie, C2FNet$^{\dagger}$, PraNet$^{\dagger}$, WSSOD, SCWSSOD, WSCOD, and SBANet) to ``Res2Net-50", and train them using the default settings.

\subsubsection{Quantitative Evaluation}

We first conduct two experiments to validate the proposed model’s learning ability on two seen datasets (\ie, Kvasir and CVC-ClinicDB). As shown in Table \ref{tab:2result}, it can be seen that our CEA-Net performs better than other weakly-supervised methods by a large margin. Besides, our SAM-CLNet outperforms CEA-Net, which indicates that SAM can effectively provide supplemental supervision signals to improve the segmentation performance. Compared to CEA-Net, our SAM-CLNet achieves $1.8\%$, $2.5\%$, and $3.4\%$ improvements over CEA-Net in terms of mDice, mIoU, and $F_\beta^\omega$ on the Kvasir dataset, respectively. This suggests that our models have a strong learning ability to accurately segment polyps in a weakly-supervised manner.

Additionally, we conduct three experiments on three unseen datasets (\ie, CVC-ColonDB, ETIS-LaribPolypDB, and CVC-300) to test the proposed model’s generalizability. As reported in Table \ref{tab:3result}, it can be observed that our CEA-Net outperforms other comparison methods by a large margin in most metrics. For example, our CEA-Net achieves $7.9\%$, $15.1\%$, $5.3\%$, and $1.9\%$ improvements over WSCOD in terms of mDice, mIoU, $S_\alpha$, and $F_\beta^\omega$ on the CVC-300 dataset, respectively. Compared to C2FNet$^{\dagger}$, our CEA-Net achieves $1.7\%$, $6.1\%$, $2.1\%$, and $1.7\%$ improvements in terms of mDice, mIoU, $S_\alpha$, and $E_\phi^{max}$ on the CVC-ColonDB dataset, repectively. More importantly, when leveraging SAM-guided collaborative
learning strategy, our SAM-CLNet further improves the segmentation performance than CEA-Net. On the ETIS-LaribPolypDB dataset, SAM-CLNet achieves $6.4\%$, $8.9\%$, $3.6\%$, $9.1\%$, and $4.2\%$ improvements over CEA-Net in terms of the first five metrics. In summary, our models has better generalization ability than other methods for weakly-supervised polyp segmentation. 


\subsubsection{Qualitative Comparison}

Visual segmentation results obtained from different methods are depicted in Fig.~\ref{fig5}. It can be observed that the results of our models are more similar to the ground truth maps, and our models perform better than all comparison methods in tackling multiple challenges. For example, the polyp illustrated in the first row is particularly small in size. Despite this, our models accurately segment this minuscule polyp while methods such as WSSOD, SCWSSOD, and WSCOD yield non-region fragments. In the $2^{nd}$ row, our CEA-Net achieves a relatively better segmentation result than all other methods, while they fail to locate and segment complete polyp regions. Compared to CEA-Net, our SAM-CLNet further enhances the segmentation performance and yields a satisfactory result. In the $3^{rd}$ and $4^{th}$ rows, the polyps are considerably large in size, which introduces a challenge for accurately segmenting the entire polyp region. In this case, our models excel over comparison methods, while WSSOD, SCWSSOD, SBA-Net, C2FNet$^{\dagger}$, and PraNet$^{\dagger}$ fail to locate several polyp regions. In the $5^{th}$ row, a polyp is visually embedded in its surrounding mucosa. The indistinct boundaries between the polyps and background pose a significant challenge for segmentation. Even with this obstacle, our methods outperform others by segmenting the polyp more accurately. Moreover, our methods effectively segment multiple polyps and exhibit an advantage in locating fine boundaries, as illustrated in the $6^{th}$ row. Overall, the qualitative comparison results further affirm that the proposed models (CEA-Net and SAM-CLNet) exhibit robust performance in handling various challenging factors for weakly-supervised polyp segmentation.

\begin{table}[t!]
 	\centering
	\renewcommand{\arraystretch}{1.2}
	\setlength\tabcolsep{1.0pt}
	\caption{Ablation study for our CEA-Net. 
	}\label{tab:ablation1}\vspace{-0.15cm}
	\begin{tabular}{c|ccc|ccc|ccc|ccc}
		\hline
    \multirow{2}{*}{No.} & \multicolumn{3}{c|}{Settings} & \multicolumn{3}{c|}{CVC-ClinicDB} & \multicolumn{3}{c|}{CVC-300} & \multicolumn{3}{c}{Kvasir}  \\
  \cline{2-13} & Base &CEM &FAM  &mDice &mIoU &$S_{\alpha}$ &mDice &mIoU &$S_{\alpha}$ &mDice &mIoU &$S_{\alpha}$ \\
    \hline
    {1} &{\checkmark}  &  &                    & .762 & .712 & .838 & .715 & .663 & .821 & .778 & .641 & .806 \\
    {2}& \checkmark  & \checkmark &            & .820 & .732 & .885 & .798 & .702 & .887 & .830 & .685 & .867  \\
    {3}& \checkmark  &  & \checkmark           & .829 & .737 & .884 & .832 & .744 & .903 & .834 & .688 & .871 \\
    {4}& \checkmark  & \checkmark & \checkmark & \textbf{.845} & \textbf{.781} & \textbf{.891} & \textbf{.866} & \textbf{.791} & \textbf{.915} & \textbf{.842} & \textbf{.769} & \textbf{.881} \\
    \hline
	\end{tabular}\vspace{-0.15cm}
\end{table}

\begin{table}[t!]
 	\centering
	\renewcommand{\arraystretch}{1.2}
	\setlength\tabcolsep{1.8pt}{
	\caption{Ablation study for validating the effectiveness of key strategies in SAM-guided mask generation network. 
	}\label{tab:ablation2}\vspace{-0.15cm}
	\begin{tabular}{c|ccc|ccc|ccc}
		\hline
    \multirow{2}{*}{\textbf{Methods}} & \multicolumn{3}{c|}{CVC-ClinicDB} & \multicolumn{3}{c|}{CVC-300} & \multicolumn{3}{c}{Kvasir} \\
  \cline{2-10} &mDice &mIoU &$S_{\alpha}$ &mDice &mIoU &$S_{\alpha}$ &mDice &mIoU &$S_{\alpha}$  \\
    \hline
    {$Box_{1}$} & 0.822 & 0.728 & 0.878 & 0.811 & 0.712 & 0.886 & 0.818 & 0.727& 0.856 \\
    {$Box_{2}$} & 0.821 & 0.734 & 0.873 & 0.828 & 0.764 & 0.897 & 0.824 & 0.739 & 0.857 \\
    {Mask-offline}      & 0.826 & 0.754 & 0.877 & 0.863 & 0.785 & 0.910 & 0.825 & 0.753 & 0.862  \\
    \hline
    {SAM-CLNet}& \textbf{0.853} & \textbf{0.783} & \textbf{0.897} & \textbf{0.876} & \textbf{0.800} & \textbf{0.922} & \textbf{0.857} & \textbf{0.788} & \textbf{0.882}   \\
    \hline
	\end{tabular} }\vspace{-0.25cm}
\end{table}

\subsection{Ablation Study}

We conduct a series of ablation studies to investigate the effectiveness of each key module in our CEA-Net and different strategies in SAM-CLNet. 

\subsubsection{Ablation Study on CEA-Net} To evaluate the effectiveness of each key component in the CEA-Net, we carry out four ablation experiments, as shown in \tabref{tab:ablation1}). In the No.1 experiment, we remove all CEMs and FAMs, while only using a concatenation to integrate the related features (denoted as ``Base"). In the No.2 experiment, we remove the FAM and utilize a concatenation operation to combine the feature from the previous layer of the decoder with the feature from the encoder. In the No.3 experiment, we remove CEM and utilize a simple concatenation operation to integrate the adjacent features. Finally, No.4 is the full version of our CEA-Net. 

$\bullet$ \textbf{Effectiveness of CEM}: We investigate the importance of the Cross-level Enhancement Module (CEM). From \tabref{tab:ablation1}, it can be observed that No.2 (Base + CEM) outperforms No.1, indicating that our CEM can fully fuse the cross-level features to enhance the feature representations' ability, resulting in an improved segmentation performance. 

$\bullet$ \textbf{Effectiveness of FAM}: From the results reported in \tabref{tab:ablation1}, No.3 performs better than No.1. This indicates that introducing the FAM can enable our model to locate and segment polyps more accurately. Moreover, comparing No.4 with No.2, the use of FAM can also improve the segmentation performance, further suggesting the effectiveness of the proposed FAM.


\subsubsection{Ablation Study on SAM-guided Mask Generation}

\begin{figure}[!t]
  \centering
  \footnotesize
  \includegraphics[width=0.5\textwidth]{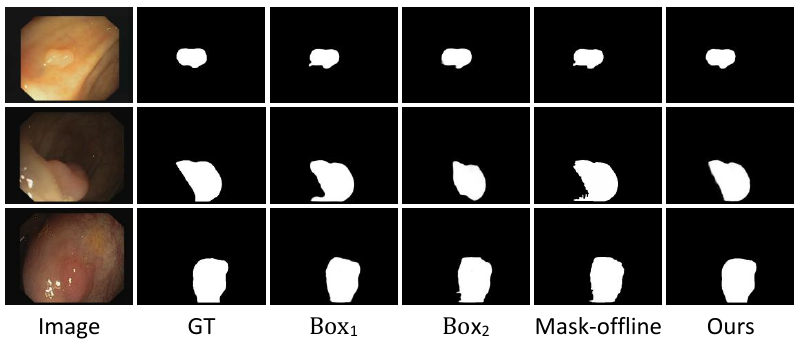}\vspace{-0.25cm}
  \caption{\footnotesize Visual segmentation results of box-prompt, offline-mask, and our online-mask.}
  \label{box_visual}\vspace{-0.2cm}
\end{figure}

To investigate the effectiveness of the prompt generation strategy, we conduct two ablation experiments using two individual bounding boxes as prompts, respectively. As presented in Table \ref{tab:ablation2}, it is evident that when $Box_1$ is used as the prompt, there is a noticeable decline in performance across multiple metrics such as mDice, mIoU, and $S_{\alpha}$ for all datasets. Furthermore, when solely relying on $Box_2$, substantial declines in these three metrics can also be observed, particularly on the CVC-ClinicDB dataset. Besides, we adopt an off-line strategy to obtain the SAM-guided masks (denoted as ``Mask offline"), in which the masks are not updated during training. Compared to ``Mask offline", our collaborative learning strategy achieves the best performance through online fine-tuning of SAM, effectively updating the SAM-guided prompts. Moreover, Fig.~\ref{box_visual} displays the visual segmentation results obtained by employing various prompts and updating strategies. The results reveal that our SAM-CLNet consistently achieves highly accurate segmentation maps, characterized by finely delineated boundaries and intricate details.




\begin{figure}[!t]
  \centering
  \footnotesize
\includegraphics[width=0.5\textwidth]{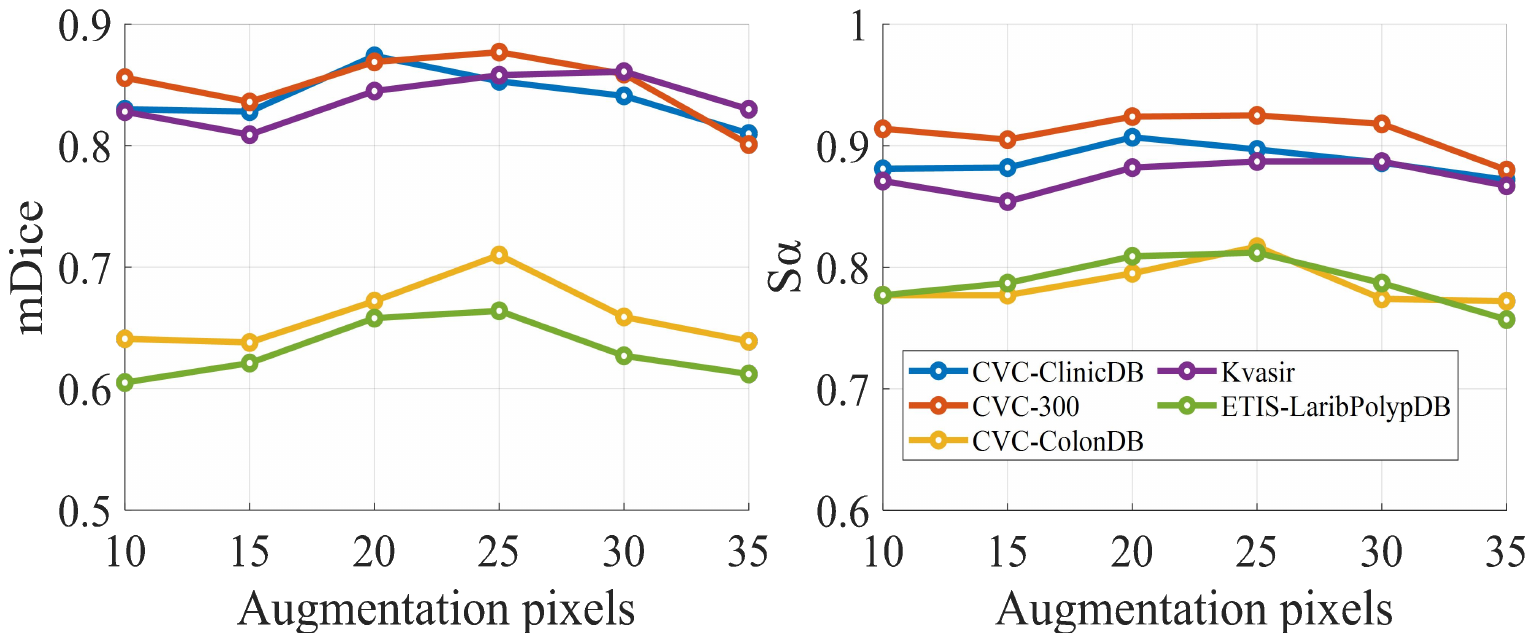}\vspace{-0.3cm}
  \caption{\footnotesize Effects of different augmentation pixel values. 
  }
  \label{augmentation}\vspace{-0.25cm}
\end{figure}

Moreover, it is worth noting that our SAM-CLNet employs a collaborative learning framework to fine-tune SAM online. In order to verify the efficacy of this online fine-tuning strategy, we conduct an experiment (\ie, Mask-offline) where we obtain the SAM-guided masks prior to training, and SAM is not engaged in the training process. As shown in Table \ref{tab:ablation2}, our model, which utilizes an online fine-tuning strategy, outperforms the one using an offline approach in terms of performance. This provides further evidence for the effectiveness of the online collaborative learning framework.


Finally, we investigate the impacts of different augmentation pixel values on the segmentation results. Fig.~\ref{augmentation} illustrates the effects in two metrics (\ie, mDice and $S_{\alpha}$) across five datasets in response to six different pixel values. Based on the results, it can be seen that there is minimal variation in performance when using different values. However, it is noteworthy that our model achieves relatively superior segmentation performance when the augmentation pixel value is set to $20$ or $25$. Thus, we have chosen a value of $25$ for augmentation pixels.

\section{Conclusion}

In this paper, we have proposed a novel SAM-CLNet framework for weakly-supervised polyp segmentation, which effectively utilizes the strength of SAM to boost the segmentation performance. Within SAM-CLNet, we propose a CEA-Net as the basic model to generate segmentation maps. We use the CEM to integrate the adjacent features in the encoder. Then, the FAM is used to aggregate the outputs from the previous CEMs, and then the aggregated feature is combined with the encoder's feature to generate the final segmentation results. Moreover, we utilize the segmentation maps and scribble annotations to form precise prompts, which are passed through SAM to generate SAM-guided masks. The SAM-guided masks can provide supplemental supervision signals to enhance the model training. Further, we design a box-augmentation strategy and an image-level Filtering Mechanism to enhance the prompts. Experimental results on scribble-annotated polyp datasets demonstrate the superiority of our SAM-CLNet over other methods.

\ifCLASSOPTIONcaptionsoff
  \newpage
\fi

\bibliographystyle{IEEEtran}
\bibliography{release_tmi}

\end{document}